\documentclass{article}

\usepackage{arxiv}
\usepackage{wrapfig}
\usepackage[utf8]{inputenc} 
\usepackage[T1]{fontenc}    
\usepackage{hyperref}       
\usepackage{url}            
\usepackage{booktabs}       
\usepackage{amsfonts}       
\usepackage{nicefrac}       
\usepackage{microtype}      
\usepackage{lipsum}		
\usepackage{graphicx}
\usepackage{natbib}
\usepackage{doi}
\usepackage{float}
\usepackage{caption}
\usepackage{subcaption}
\usepackage{booktabs,chemformula}
\usepackage{multirow}

\title{Improving Knee Joint Angle Prediction through Dynamic Contextual Focus and Gated Linear Units}

\author{$^a$Lyes Saad Saoud, ~$^b$Humaid Ibrahim, ~$^c$Ahmad Aljarah, ~$^{a,c,*}$Irfan Hussain, \\
	$^a$Khalifa University Center for Autonomous and Robotic Systems, Khalifa University, \\
	$^b$National Service and Reserve Authority, Khalifa University, \\
 $^c$Advanced Research and Innovation Center, Khalifa University, \\Abu Dhabi, United Arab Emirates, P O Box 127788, Abu Dhabi, UAE \\
        \texttt{lyes.saoud@ku.ac.ae, ahmad.aljarah@ku.ac.ae, irfan.hussain@ku.ac.ae ($^*$Corr.)}
}

\renewcommand{\undertitle}{Under consideration at Pattern Recognition Letters}
\renewcommand{\shorttitle}{Improving Knee Joint Angle Prediction through Dynamic Contextual Focus and Gated Linear Units}

\hypersetup{
pdftitle={Improving Knee Joint Angle Prediction through Dynamic Contextual Focus and Gated Linear Units},
pdfsubject={q-bio.NC, q-bio.QM},
pdfauthor={Lyes Saad Saoud, Humaid Ibrahim, Ahmad Aljarah, Irfan Hussain},
pdfkeywords={Knee Joint Angle Prediction, Deep Learning, Multi-step Forecasting, Biomechanical Analysis, Rehabilitation},
}

\begin{document}
\maketitle



\begin{abstract}

Accurate knee joint angle prediction is crucial for biomechanical analysis and rehabilitation. In this study, we introduce FocalGatedNet, a novel deep learning model that incorporates Dynamic Contextual Focus (DCF) Attention and Gated Linear Units (GLU) to enhance feature dependencies and interactions. Our model is evaluated on a large-scale dataset and compared to established models in multi-step gait trajectory prediction.
Our results reveal that FocalGatedNet outperforms existing models for long-term prediction lengths (60 ms, 80 ms, and 100 ms), demonstrating significant improvements in Mean Absolute Error (MAE), Root Mean Square Error (RMSE) and Mean Absolute Percentage Error (MAPE). Specifically for the case of 80 ms, FocalGatedNet achieves a notable MAE reduction of up to 24\%, RMSE reduction of up to 14\%, and MAPE reduction of up to 36\% when compared to Transformer, highlighting its effectiveness in capturing complex knee joint angle patterns.
Moreover, FocalGatedNet maintains a lower computational load than most equivalent deep learning models, making it an efficient choice for real-time biomechanical analysis and rehabilitation applications.

\end{abstract}




\section{INTRODUCTION}

Wearable exoskeletons find diverse applications, including healthcare, industry, space, and the military \citep{A1, A2, A3, A4}. In healthcare, they are instrumental in rehabilitating post-stroke patients and addressing age-related disorders. These exoskeletons enhance patient mobility, assist physical therapists, and reduce rehabilitation clinical costs \citep{A5}. 

An exoskeleton's performance hinges on its control strategy, governing operation and user interaction \citep{A6}. Strategies cover low, middle, and high-level control, encompassing tasks like user intention detection, terrain identification, and event estimation \citep{A7, A8}
\begin{figure*}[ht]
\centering
\includegraphics[width=0.9\textwidth]{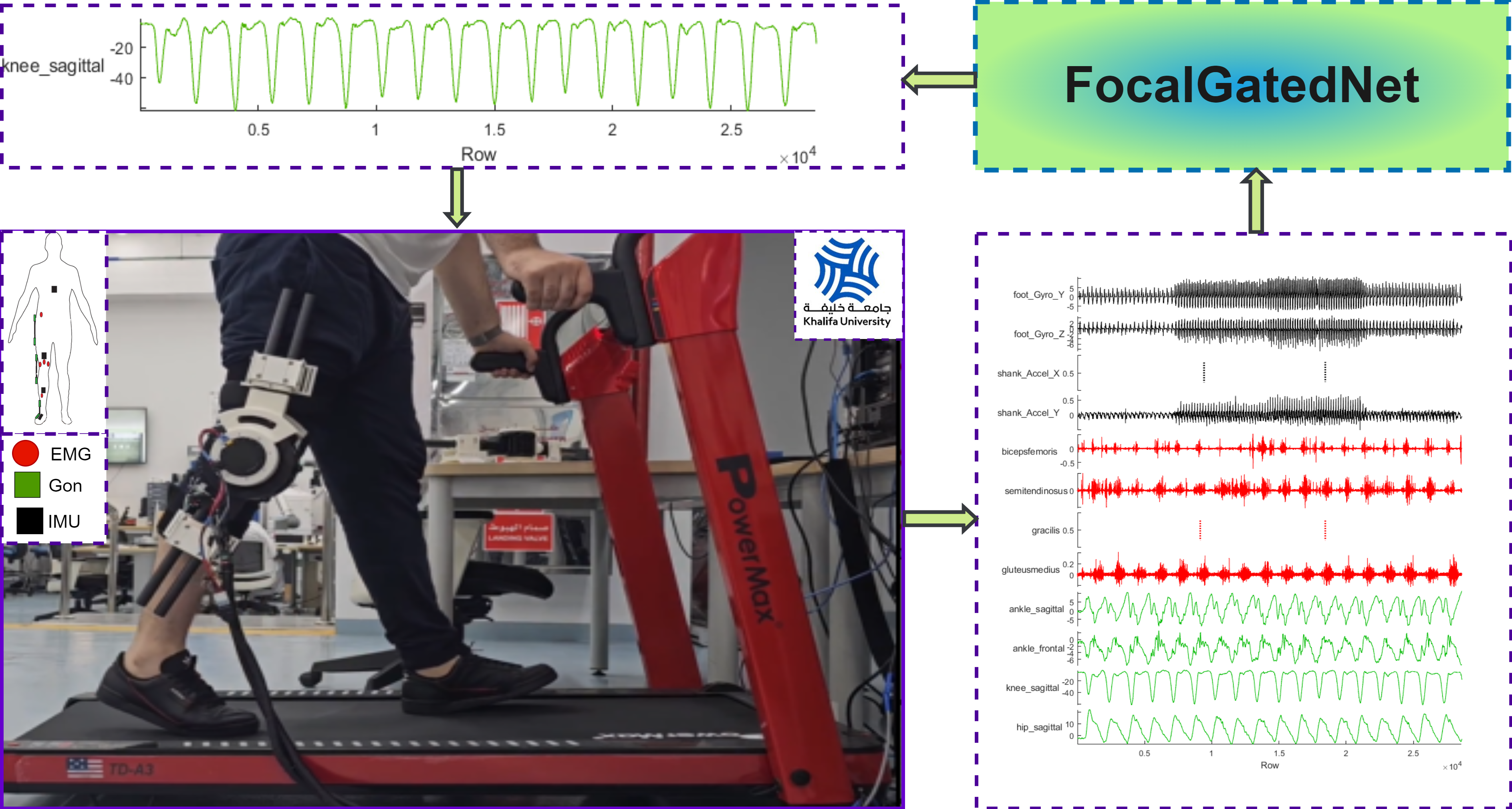}
\caption{Overview of the proposed FocalGatedNet model for long-term gait trajectory prediction in exoskeleton control systems. FocalGatedNet, a novel deep learning model, incorporates Dynamic Contextual Focus (DCF) Attention and Gated Linear Units (GLU) to augment feature dependencies and interactions. The input-output data obtained from sensors: EMG, acceleration, and
gyroscope information from 6-axis IMUs, and joint angle data from electrogoniometers. The output is a predicted long-term gait trajectory for exoskeleton control and rehabilitation applications.}  
\label{main_model}
\end{figure*}
The accuracy of the high-level controller significantly impacts the overall functionality of the exoskeleton. Typically, gait data analysis is employed to detect user intentions or estimate essential gait parameters like angular positions, velocities, and accelerations, which are essential for developing an accurate controller that improves exoskeleton performance \citep{A9}. Various sensors, including inertial measurement units (IMUs), motion capture systems, foot pressure insoles, gyroscopes, accelerometers, Electromyography (EMG), and Electroencephalography (EEG), are used to estimate kinetic and kinematic parameters, muscle activities, and brain functions. Researchers have developed multiple algorithms to analyze and process sensor data, infer human intentions, and predict essential gait parameters.

Recent advancements in machine learning and deep learning have transformed gait analysis, enabling automatic and faster assessments for more responsive low-level controllers \citep{A7}. Deep learning, in particular, has outperformed traditional methods like Support Vector Machines (SVM) and Multilayer Perceptron (MLP) in analyzing lower-limb locomotion and pathological gait \citep{A10, A11, A17}.
Despite the effectiveness of these deep learning models, particularly Transformers, their suitability for long-term gait trajectory prediction in exoskeleton control systems remains unexplored.

Exoskeletons inherently exhibit mechanical delays, posing a challenge to their control systems, which can affect response times. In applications such as rehabilitation, where exoskeletons move in tandem with users, accurate and timely responses are crucial for achieving desired outcomes, such as assisting in movement and reducing metabolic costs. Utilizing a feed-forward input that contains future gait trajectories can mitigate this issue. While there have been numerous studies on deep learning models for gait prediction, they typically employ small output window sizes consisting of a few time steps. To our knowledge, the long-term performance of deep learning models in gait prediction has yet to be thoroughly investigated.

To address these limitations, this paper introduces a novel deep learning model, FocalGatedNet, and evaluates its performance, alongside other deep learning models, across varying output lengths. FocalGatedNet, based on the Transformer model, incorporates Dynamic Contextual Focus (DCF) Attention and Gated Linear Units (GLU) to enhance feature dependencies and interactions, resulting in accurate knee joint angle prediction.

We evaluate the proposed FocalGatedNet model using a publicly available dataset comprising gait data from both healthy individuals and those with pathological conditions. Our experimental results demonstrate that the proposed model outperforms state-of-the-art deep learning models in knee joint angle prediction accuracy. The inclusion of DCF in the FocalGatedNet model effectively addresses the sample imbalance issue in the dataset, enhancing the model's focus on challenging samples. Additionally, the GLU layer in our proposed model captures long-term dependencies in gait data, further enhancing knee joint angle prediction accuracy.

The main contributions of this paper can be summarized as follows:
\begin{itemize}
  \item Introduction of FocalGatedNet: We propose FocalGatedNet, a novel deep-learning model for predicting knee joint angles in lower-limb exoskeleton control. Our model integrates two key components, namely Dynamic Contextual Focus (DCF) Attention and Gated Linear Units (GLU), to effectively capture feature dependencies and interactions, thereby improving knee joint angle prediction accuracy.

 \item Evaluation on a Public Dataset: We evaluate our proposed model using a publicly available dataset containing time-series gait data from healthy individuals. Our results demonstrate the model's effectiveness in accurately predicting knee joint angles for various output window sizes.
\end{itemize}
The remainder of this paper is organized as follows: Section II provides a brief overview of related work in gait analysis and knee joint angle prediction. Section III introduces the proposed FocalGatedNet model architecture, explaining DCF Attention and GLU in detail. In Section IV, we describe the experimental setup and dataset used to evaluate the model's performance. Section V presents and discusses the experimental results, along with potential applications of our proposed model in wearable robotics and exoskeletons. Finally, Section VI offers concluding remarks and outlines future research directions.

\section{Related works}

\subsection{Long-term series forecasting}
In the field of exoskeleton control, deep learning models have gained prominence for predicting trajectories as inputs to low-level controllers. Models like LSTM and CNN are essential for predicting parameters like joint angles \cite{LSTM_Angle, LSTM_CNN_Angle}, accelerations, and velocities \cite{LSTM_acceleration1, LSTM_acceleration2}. However, these studies often focus on short-term predictions.


In the realm of long-term time series forecasting (LTSF), innovative models have emerged. Convolutional Neural Networks (CNNs), including the Temporal Convolutional Network (TCN) \cite{TCN}, have been developed. B. Fang et al. achieved competitive results using TCNs for gait recognition, albeit limited to one-step predictions.
Another noteworthy model is the Gated Convolutional Neural Network (Gated CNN) \cite{GatedCNN}, addressing long-term dependencies in CNNs. The Gated Linear Unit (GLU) in Gated CNNs selectively filters information from previous time steps, enhancing feature selection. GLU offers distinct advantages over other activation functions, capturing long-range dependencies and producing sparser representations, beneficial for extensive sensor data in gait datasets. In our recent work, we introduced TempoNet \citep{saoud_tempoNet_2023}, a Transformer-based model equipped with dynamic temporal attention, which has shown promising results. 

Transformer models \cite{vaswani, Gait_Prediction_Transformer} have made strides in LTSF, excelling in processing sequential data efficiently. Traditional models like RNNs and LSTMs are limited by their sequential processing nature, especially for long sequences.
Transformers identify essential patterns and learn dependencies across time steps. However, they face challenges due to quadratic time and memory complexity ($\mathcal{O}(L^2)$) for longer sequences.

To overcome this, models like Informer \cite{informer} introduced ProbSparse attention ($\mathcal{O}(L \log L)$ complexity). Autoformer \cite{autoformer} introduced the Auto-Correlation mechanism, capturing correlations between time steps efficiently.

Non-transformer models like DLinear and NLinear \cite{LTSF} employ preprocessing techniques, relying on linear layers. DLinear uses a scheme similar to Autoformer, while NLinear performs well in cases with data distribution shifts, offering linear complexity ($\mathcal{O}(L)$).

In our study, we assess the performance of our proposed FocalGatedNet in comparison to Transformer and non-Transformer models across various output lengths for exoskeleton control \cite{informer, autoformer, LTSF}.

\subsection{Exoskeleton Delay}

The concept of compensating for exoskeleton transmission delay through future value prediction was explored in \citep{exoskeleton_delay_prediction}. They predicted the trajectory of 10 time steps (approximately 200 ms) using two models, MLP and CNN, with the goal of mitigating the response time of the exoskeleton control system. Fang et al. \citep{exoskeleton_mechanical_delay} highlighted that delays stemming from various aspects of an exoskeleton's mechanical structure can significantly reduce the actual assistance provided by the exoskeleton, impacting simulated energy savings.

Accurately identifying the current phase of the gait cycle can enhance assistive-powered prosthesis control, providing critical information for determining appropriate angle, angular velocity, and torque values. This improved control can lead to reduced energy expenditure while walking with a powered limb, benefiting the patient.

While predicting up to a certain point in the gait cycle can be advantageous for exoskeleton control, striking a balance between model prediction error and output length can be challenging. Yi et al. \citep{EMD_prediction} conducted an experimental study and determined that optimal output sizes for knee joint angle prediction fell within the interval of 27 ms to 108 ms. This interval was influenced by the concept of electromechanical delay (EMD) \citep{viitasalo}, which posits that EMG signals exhibit a delay of 30-100 ms between muscle activation (as detected by the EMG sensor) and the resulting force and movement generated by the muscle.

Their findings supported this interval, with model performance declining significantly when predicting beyond the 30-100 ms range. However, their study did not evaluate prediction times below the lower limit of this interval. In our research, we will investigate this same prediction interval while also including a one-step prediction time.

Our primary objective in this study is to utilize the unique features of EMD along with our proposed model to investigate potential relationships between EMD and prediction time. In addition, we aim to maximize exoskeleton transmission delay compensation through longer prediction lengths. To achieve this, we leverage EMG and IMU data as our primary features. By doing so, we hope to contribute to the development of more effective assistive-powered exoskeletons/prostheses that can provide patients with better mobility and reduce their energy expenditure while walking with a powered limb.

\section{FocalGatedNet}

\begin{figure}[H]
\centering
\includegraphics[width=0.5\textwidth]{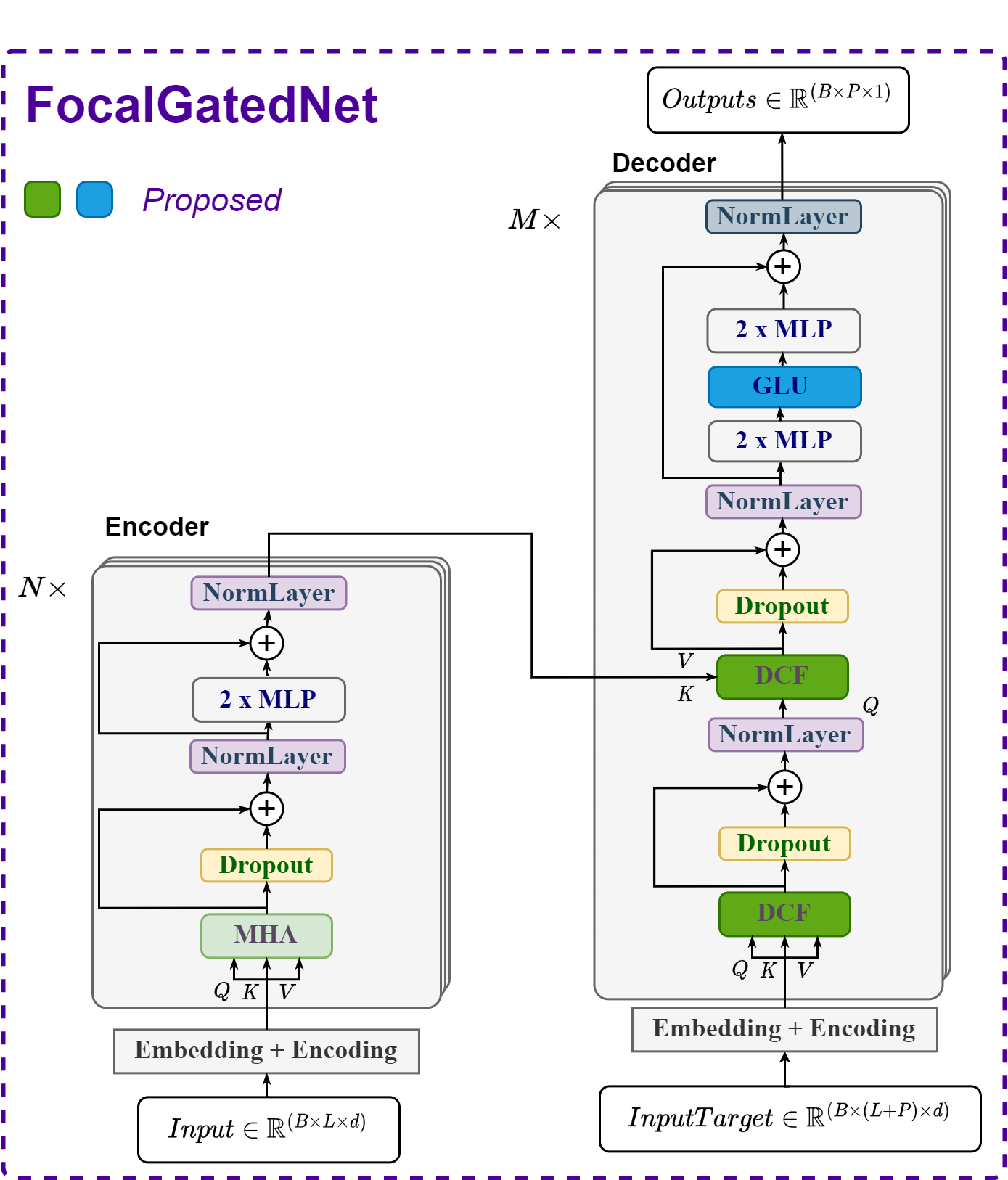}
\caption{FocalGatedNet Structure. The FocalGatedNet consists of N encoders and M decoders. The encoder block is the same as the original Transformer model. The decoder has two DCF attention blocks in place of Multi-Head Attention and a GLU block with an additional feed-forward network.}  
\label{model_architecture}
\end{figure}

FocalGatedNet enhances the core Transformer by introducing a modified decoder that integrates stacked Dynamic Contextual Focus (DCF) attention layers and a Gated Linear Unit (GLU), as illustrated in Figure \ref{model_architecture}.

\begin{figure}[H]
\centering
\includegraphics[width=0.25\textwidth]{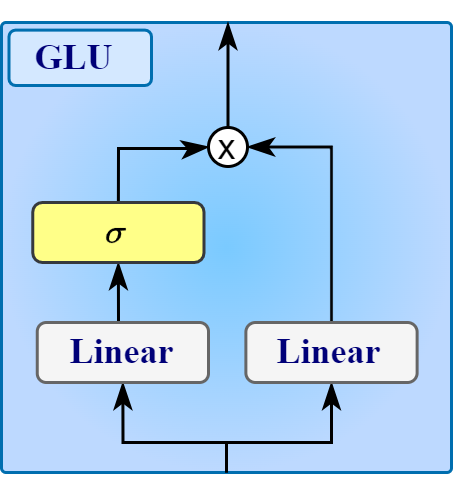}
\includegraphics[width=0.25\textwidth]{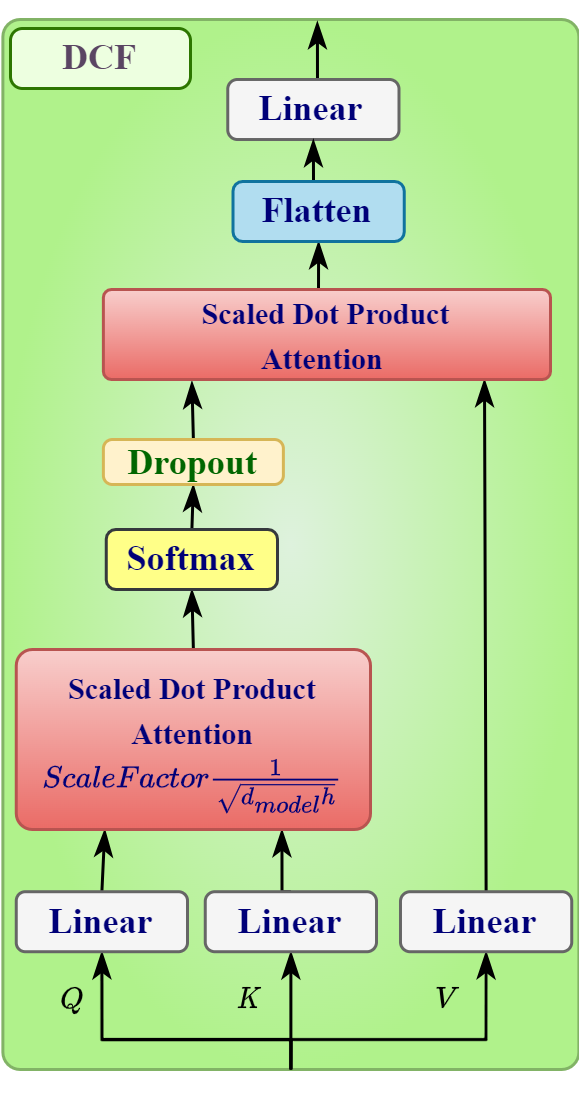}
\caption{The proposed GLU and DCF Architectures.}
\label{GLU}
\end{figure}

The primary distinction lies in how DCF Attention and the conventional Attention mechanism \citep{vaswani} interact with the input sequence. DCF Attention adopts a hierarchical approach, while the traditional Attention treats all positions within the input sequence uniformly. The architectural differences of DCF are visually depicted in Figure \ref{GLU}.

In mathematical terms, DCF operates as follows:

Given an input sequence $x\in\mathbb{R}^{B\times L\times d_{model}}$, where $B$ represents the batch size, $L$ denotes the sequence length, and $d_{model}$ signifies the hidden neurons of the input embedding, the self-attention mechanism computes the query, key, and value representations as follows:

\begin{equation} \label{eq1}
\begin{split}
Q = x \; W_Q &\in \mathbb{R}^{B \text{x} L \text{x} d_{model}} \\ 
K = x \; W_K &\in \mathbb{R}^{B \text{x} L \text{x} d_{model}} \\
V = x \; W_V &\in \mathbb{R}^{B \text{x} L \text{x} d_{model}} 
\end{split}
\end{equation}
where $\text{W}_\text{Q}, \text{W}_\text{K}, \text{W}_\text{V}$ are linear projections, each of size $d_{model}\times d_{model}$.

Then, the dot-product attention scores between queries and keys are computed as:
\begin{equation} \label{eq2}
    A = Softmax \left(\frac{QK^T}{\sqrt{d_{model}h}}\right)
\end{equation}
where $h$ is the number of attention heads, $A$ is the attention weights tensor, and $\frac{1}{\sqrt{d_{model}h}}$ is the scaling factor SF.

If a mask is provided, it is applied to the attention scores as:
\begin{equation} \label{eq3}
    A = A \odot M
\end{equation}
where $M\in {0,1}^{B\times h\times L\times L}$ is the mask tensor.

Next, the contextual focus vector is computed as:
\begin{equation} \label{eq4}
    C = \sum_{i=1}^{L} A_i \cdot V_i \in \mathbb{R}^{B \text{x} h \text{x} L \text{x} L }
\end{equation} 

The contextual focus vector is passed through a Softmax function with dropout to compute the attention weights:
\begin{equation} \label{eq5}
    W = Dropout \left( Softmax \left( Flatten (C) \right)   \right) \in \mathbb{R}^{B \text{x} h \text{x} L}
\end{equation}
where $\text{flatten}(C)$ denotes reshaping $C$ from $\mathbb{R}^{B\times h\times d_{\text{model}}}$ to $\mathbb{R}^{Bh\times d_{\text{model}}}$.

Finally, the weighted sum of the attention weights and values are computed as:
\begin{equation} \label{eq6}
    O = cat(W, V) \; W^O \in \mathbb{R}^{B \text{x} L \text{x} d_{model}}
\end{equation}
where $\text{cat}(W, V)$ denotes concatenating $W$ and $V$ along the attention head dimension, and $\text{out}$ is another linear projection layer of size $d_{\text{model}}\times d_{\text{model}}$.

The DCF uses a different parameterization of the attention weights, which is more effective than the traditional dot product attention used in the original Attention mechanism. 
We then incorporate a gating mechanism, the GLU, as seen in Fig. \ref{GLU}. This mechanism can help the model selectively attend to different parts of the input sequence and further improve the quality of the learned representations.

The GLU output can be computed as follows:
\begin{equation} \label{eq}
GLU(x) = \sigma(W_g * x) \odot (W_h*x)
\end{equation}
where $\sigma$ is the sigmoid activation function,  $\odot$ represents element-wise multiplication, $x$ is the input to the layer, $W_g$ and $W_h$ are the weights of the gating and linear layers, respectively. The GLU has a complexity of $\mathcal{O}(L/k)$, where $k$ is the kernel width. 

The incorporation of GLU diminishes input dimensionality through feature gating, enabling the model to concentrate on pertinent features. This reduction, akin to Gated CNN, curtails the likelihood of overfitting, augments generalization, and leads to quicker inference times. This swiftness holds particular significance for exoskeleton control, as it facilitates compensation for transmission delays while preserving prediction precision. The impacts of GLU and DCF are examined in our ablation study in Section \ref{results}.

\section{Results and Discussion}
\label{results}
This study employs an open-source dataset that encompasses human gait kinematics and kinetics during various locomotion behaviors \citep{dataset}. The dataset includes recordings from 22 able-bodied subjects performing a wide range of locomotion actions, such as walking on level ground at varying paces, both clockwise and counterclockwise circuits, treadmill walking at multiple speeds, as well as ascending and descending ramps with different inclines and stairs with varying step heights. The dataset provides three primary time-series sensor signals:

\textbf{EMG Data:} Collected at a rate of 1000 Hz and bandpass filtered (20 Hz - 400 Hz), this data includes readings from 11 muscle groups, including gluteus medius, rectus femoris, and tibialis anterior.

\textbf{IMU Data:} Recorded at a frequency of 200 Hz and lowpass filtered at 100 Hz, this data originates from 4 IMUs placed on the torso, thigh, shank, and foot segments.

\textbf{Goniometer Data (GON):} Collected at 1000 Hz and subjected to lowpass filtering (20 Hz cutoff frequency), this data comes from 3 GON sensors positioned on the hip, knee, and ankle.

In addition to raw sensor data, the dataset offers processed biomechanics data, covering aspects like inverse kinematics/dynamics, joint power, gait cycles, force plates, and motion capture data.

The input features consisted of 40 sensors (EMG, GON, and IMU), with knee sagittal angle as the output. Data was normalized to have a mean of zero and a standard deviation of one. The data interpolation aligned IMU and toe/heel events with the EMG and goniometer data sampling rates. The lookback window was set to 128 ms, and forecasting horizons ranged from 1 ms to 100 ms, addressing muscle activation delays and exoskeleton mechanical responses.

This seamless integration of data preprocessing steps enabled us to effectively harness the wealth of information contained within the dataset for accurate knee sagittal angle predictions.

All models were implemented in PyTorch 1.13.1 and trained on a system equipped with an NVIDIA GeForce RTX 4090 Ti GPU. We utilized the Adam optimizer with an adaptive learning rate starting at $10^{-4}$ and decaying by a factor of 2 per epoch. The training ran for 10 epochs with early stopping after 3 epochs of no improvement. A batch size of 32 was employed, and 10 iterations were performed, selecting the best results. Data from a single subject was divided into an 80\% training set and a 20\% testing set.


FocalGatedNet featured 3 encoder and 2 decoder layers. Remarkably, omitting temporal embeddings improved its performance. Other transformer-based models often performed better without positional embeddings. We maintained attention heads at 8, with model dimensions of 512 and 2048 for the feed-forward network.

In our study, we conducted a comprehensive evaluation of FocalGatedNet's performance in comparison to other recent models, including LSTM \citep{HochSchm97}, GRU \citep{10.1145/3377713.3377722}, Transformer \citep{vaswani}, Autoformer \citep{autoformer}, Informer \citep{informer}, DLinear, and NLinear \citep{LTSF}. This assessment was based on key metrics, namely MAE, RMSE, MAPE, and R². Furthermore, we conducted an ablation study to assess the impacts of the DCF and GLU blocks. For shorter prediction horizons, the GLU-only model outperforms the base transformer and the GLU+DCF model. However, for horizons of 60 ms and beyond, FocalGatedNet excels, emphasizing the critical role of the GLU block for overall model performance, as the DCF's performance degrades without it.

\begin{table}[ht]
\caption{Ablation study of the Gated Linear Unit and Dynamic Contextual Focus blocks. The best results are highlighted in bold.}
\label{ablation}
\centering
\small 

\begin{tabular}{p{0.85cm}p{0.85cm}p{0.85cm}p{0.85cm}p{0.85cm}p{0.85cm}p{0.85cm}} 
\hline

\multirow{2}{*}{Model}
& \multicolumn{2}{p{1.7cm}}{GLU+DCF}                       
& \multicolumn{2}{p{1.7cm}}{DCF}                      
& \multicolumn{2}{p{1.7cm}}{GLU} \\ 
\cline{2-7}

{ }
& \multicolumn{1}{p{0.85cm}}{MAE} & \multicolumn{1}{p{0.85cm}}{RMSE}    
& \multicolumn{1}{p{0.85cm}}{MAE} & \multicolumn{1}{p{0.85cm}}{RMSE}    
& \multicolumn{1}{p{0.85cm}}{MAE} & \multicolumn{1}{p{0.85cm}}{RMSE}  \\ 
\hline

1
& 0.587 & 0.681
& 0.496 & 0.598
& \textbf{0.280} & \textbf{0.355} \\

20
& \textbf{0.681} & \textbf{0.837}   
& 6.394 & 7.024
& 0.690 & 0.849 \\

40
& 0.985 & 1.220   
& 6.992 & 7.865
& \textbf{0.945} & \textbf{1.176} \\

60
& \textbf{1.058} & \textbf{1.404}   
& 7.832 & 8.900
& 1.125 & 1.412 \\

80
& \textbf{1.115} & \textbf{1.584}   
& 12.361 & 13.827
& 1.507 & 1.920 \\

100
& \textbf{1.234} & \textbf{1.793} 
& 13.949 & 15.235
& 1.445 & 2.002 \\
\hline

\end{tabular}
\end{table}
\begin{table}[H]

\centering
\caption{Performance Metrics for 1 ms Ahead Forecasting}
\label{table11}
\small 
\begin{tabular}{p{2.7cm} p{1cm} p{1cm} p{1cm} p{1cm}} 
\hline
\textbf{Models} & \textbf{MAE} & \textbf{RMSE} & \textbf{MAPE} & \textbf{R²} \\
\hline
DLinear  & 0.225 & 0.250 & 0.030 & 99.99 \\
NLinear & \textbf{0.147} & \textbf{0.203} & \textbf{0.010} & \textbf{99.99} \\
LSTM  & 1.168 & 1.528 & 0.164 & 99.45 \\
GRU  & 1.188 & 1.557 & 0.167 & 99.43 \\
Autoformer  & 1.548 & 2.506 & 0.138 & 98.52 \\
Transformer & 0.288 & 0.370 & 0.035 & 99.97 \\
Informer & 0.608 & 0.781 & 0.059 & 99.86 \\
\textbf{FocalGatedNet} & 0.587 & 0.681 & 0.075 & 99.89 \\
\hline
\end{tabular}
\end{table}
\begin{figure}[ht]
\centering
\includegraphics[width=0.75\textwidth]{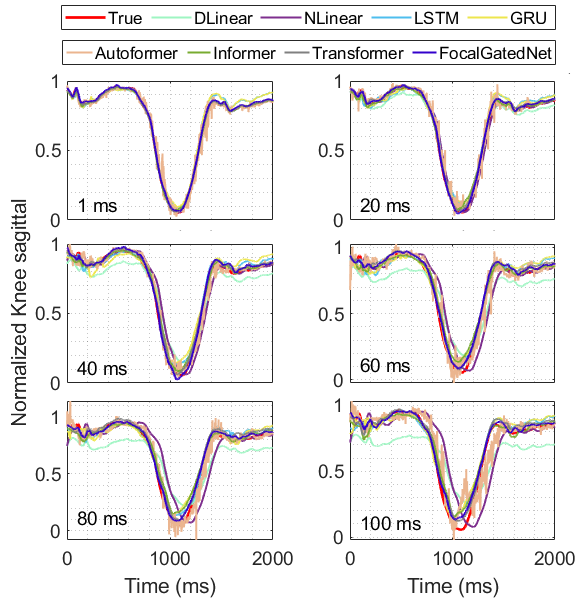}
\caption{Knee Joint Angle Predictions for Various Time Durations.
This figure presents the predicted knee joint angles for different time durations (1 ms, 20 ms, 40 ms, 60 ms, 80 ms, and 100 ms) using various deep learning models (DLinear, NLinear, LSTM, GRU,  Autoformer, Informer, FocalGatedNet). The true knee joint angle (labeled as "True") is also included for reference. Each subplot corresponds to a specific time duration, showcasing the accuracy and performance of the models in predicting knee joint angles. The legend displays the model names, and the x and y-axis labels provide additional context for interpretation.}
\label{compare}
\end{figure}


\begin{table}[H]

\centering
\small 
\caption{Performance Metrics for 20 ms Forecasting}
\label{table2}
\begin{tabular}{p{2.7cm} p{1cm} p{1cm} p{1cm} p{1cm}} 
\hline
\textbf{Models} & \textbf{MAE} & \textbf{RMSE} & \textbf{MAPE} & \textbf{R²} \\
\hline
DLinear & 3.144 & 3.518 & 0.423 & 97.09 \\
NLinear & 2.438 & 3.408 & 0.164 & 97.27 \\
LSTM  & 1.498 & 1.941 & 0.191 & 99.11 \\
GRU   & 1.528 & 1.984 & 0.177 & 99.08 \\
Autoformer   & 1.800 & 2.678 & 0.187 & 98.32 \\
Transformer  & 0.705 & 0.889 & 0.093 & 99.81 \\
Informer   & 0.821 & 1.089 & 0.116 & 99.72 \\
\textbf{FocalGatedNet} & \textbf{0.681} & \textbf{0.837} & \textbf{0.085} & \textbf{99.84} \\
\hline
\end{tabular}
\end{table}

\begin{table}[H]

\centering
\small 
\caption{Performance Metrics for 40 ms Forecasting}
\label{table3}
\begin{tabular}{p{2.7cm} p{1cm} p{1cm} p{1cm} p{1cm}} 
\hline
\textbf{Models} & \textbf{MAE} & \textbf{RMSE} & \textbf{MAPE} & \textbf{R²} \\
\hline
DLinear & 5.997 & 6.674 & 0.822 & 89.55 \\
NLinear  & 4.722 & 6.674 & 0.324 & 89.55 \\
LSTM  & 1.635 & 2.269 & 0.194 & 98.79 \\
GRU   & 2.307 & 3.355 & 0.284 & 97.36 \\
Autoformer  & 2.121 & 3.005 & 0.222 & 97.88 \\
Transformer & 0.819 & 1.057 & 0.099 & 99.74 \\
Informer & 0.990 & 1.285 & 0.137 & 99.61 \\
\textbf{FocalGatedNet} & \textbf{0.985} & \textbf{1.220} & \textbf{0.120} & \textbf{99.65} \\
\hline
\end{tabular}
\end{table}

For 1 ms ahead forecasting (Table \ref{table11}), the NLinear model stands out with exceptional performance, boasting an MAE of 0.147, RMSE of 0.203, MAPE of 0.010, and an R² of 99.99\%. FocalGatedNet closely follows as the second-best model, with slightly higher MAE, RMSE, and MAPE at 0.587, 0.681, and 0.075, respectively, but it still maintains a competitive R² at 99.89\%. This showcases FocalGatedNet's commendable performance in 1 ms forecasting, with notable relative differences compared to NLinear.

Moving to a 20 ms forecasting horizon, presented in Table \ref{table2}, FocalGatedNet takes the lead, securing the top position in all metrics. With an MAE of 0.681, RMSE of 0.837, MAPE of 0.085, and an R² of 99.84\%, FocalGatedNet demonstrates significant superiority in this timeframe, with a relative improvement compared to NLinear. At longer forecasting horizons, shown in Table \ref{table3} to Table \ref{table6} of 40 ms, 60 ms, 80 ms, and even the challenging 100 ms horizon, FocalGatedNet maintains its dominance, excelling across all metrics. This consistent performance across diverse forecasting horizons underscores FocalGatedNet's reliability and competitive edge in meeting various prediction needs.
\begin{table}[H]

\centering
\small 
\caption{Performance Metrics for 60 ms Forecasting}
\label{table4}
\begin{tabular}{p{2.7cm} p{1cm} p{1cm} p{1cm} p{1cm}} 
\hline
\textbf{Models} & \textbf{MAE} & \textbf{RMSE} & \textbf{MAPE} & \textbf{R²} \\
\hline
DLinear  & 8.360 & 9.271 & 1.150 & 79.83 \\
NLinear  & 6.749 & 9.629 & 0.476 & 78.24 \\
LSTM   & 2.029 & 2.746 & 0.269 & 98.23 \\
GRU & 2.232 & 3.475 & 0.268 & 97.17 \\
Autoformer  & 2.717 & 3.815 & 0.305 & 96.58 \\
Transformer   & 1.114 & 1.442 & 0.136 & 99.51 \\
Informer   & 1.679 & 2.240 & 0.211 & 98.82 \\
\textbf{FocalGatedNet} & \textbf{1.058} & \textbf{1.404} & \textbf{0.129} & \textbf{99.54} \\
\hline
\end{tabular}
\end{table}
Fig. \ref{compare} illustrates predicted knee joint angles for various time durations, ranging from 1 ms to 100 ms. Different deep learning models, including DLinear, NLinear, LSTM, Autoformer, Informer, and FocalGatedNet, are compared against the ground truth knee joint angle data ("True").

\begin{figure}[H]
\centering
\includegraphics[width=0.6\textwidth]{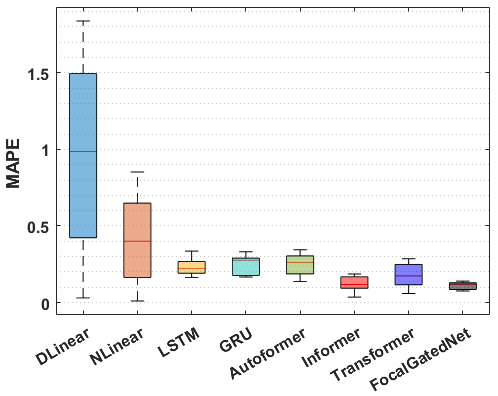}
\caption{Comparative Analysis of MAPE for Forecasting Models.}
\label{MAPE_results}
\end{figure}

Fig. \ref{MAPE_results} presents a comparative analysis of these models in terms of MAPE percentages relative to FocalGatedNet's MAPE. Each box in the plot represents a specific model, color-coded for clarity. Notably, FocalGatedNet consistently outperforms other models in MAPE after 20 ms across all forecasting horizons. For instance, at 100 ms ahead of forecasting, FocalGatedNet achieved an approximately 84\% lower MAPE than the second-best model, LSTM. These findings highlight FocalGatedNet's exceptional accuracy in forecasts across diverse horizons, positioning it as a valuable tool for precision-critical applications.

\begin{figure}[H]
\centering
\includegraphics[width=0.6\textwidth]{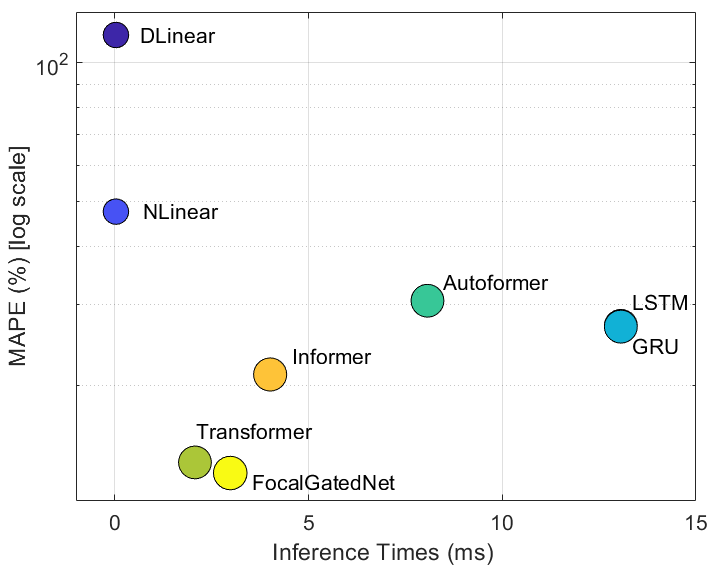}
\caption{Comparison of inference times (ms) for different machine learning models. "FocalGatedNet" demonstrates outstanding efficiency with a 2.98 ms average time.
}
\label{Performance}
\end{figure}

Fig. \ref{Performance}  presents a compelling insight into the performance of various deep learning models employed in gait analysis. Notably, FocalGatedNet stands out as the top performer, with the lowest MAPE at 12.9\%. 
\begin{table}[H]

\centering
\small 
\caption{Performance Metrics for 80 ms Forecasting}
\label{table5}
\begin{tabular}{p{2.7cm} p{1cm} p{1cm} p{1cm} p{1cm}} 
\hline
\textbf{Models} & \textbf{MAE} & \textbf{RMSE} & \textbf{MAPE} & \textbf{R²} \\
\hline
DLinear & 10.688 & 11.766 & 1.495 & 67.53 \\
NLinear & 8.772 & 12.551 & 0.649 & 63.05 \\
LSTM & 2.136 & 2.873 & 0.249 & 98.06 \\
GRU  & 2.566 & 3.516 & 0.332 & 97.10 \\
Autoformer & 2.820 & 3.874 & 0.303 & 96.48 \\
Transformer  & 1.467 & 1.849 & 0.186 & 99.20 \\
Informer  & 2.068 & 2.756 & 0.249 & 98.22 \\
\textbf{FocalGatedNet} & \textbf{1.115} & \textbf{1.584} & \textbf{0.119} & \textbf{99.41} \\
\hline
\end{tabular}
\end{table}

\begin{table}[H]

\centering
\small 
\caption{Performance Metrics for 100 ms Forecasting}
\label{table6}
\begin{tabular}{p{2.7cm} p{1cm} p{1cm} p{1cm} p{1cm}} 
\hline
\textbf{Models} & \textbf{MAE} & \textbf{RMSE} & \textbf{MAPE} & \textbf{R²} \\
\hline
DLinear & 12.973 & 14.195 & 1.839 & 52.71 \\
NLinear & 10.917 & 15.561 & 0.853 & 43.17 \\
LSTM  & 2.737 & 3.554 & 0.336 & 97.04 \\
GRU & 2.395 & 3.422 & 0.290 & 97.25 \\
Autoformer  & 3.789 & 5.207 & 0.345 & 93.64 \\
Transformer & 1.464 & 2.010 & 0.168 & 99.05 \\
Informer & 2.407 & 3.246 & 0.286 & 97.53 \\
\textbf{FocalGatedNet} & \textbf{1.234} & \textbf{1.794} & \textbf{0.140} & \textbf{99.25} \\
\hline
\end{tabular}
\end{table}

This remarkable accuracy implies that FocalGatedNet predicts knee joint angles with an error of just 0.129, highlighting its proficiency in precise gait analysis.

While FocalGatedNet shines in accuracy, it's essential to consider the trade-off between inference time and precision. For instance, models like Transformer and Autoformer exhibit outstanding accuracy as well, with MAPE values of 0.136 and 0.305, respectively. However, they demand longer inference times compared to FocalGatedNet. This figure underscores the need for a balanced approach, allowing practitioners to select the model that best aligns with their specific gait analysis requirements, whether prioritizing speed or precision.

While FocalGatingNet shows promise in knee joint angle prediction, it has limitations that require further investigation.



\begin{itemize}
  \item Transformers, like FocalGatingNet, excel with longer sequential data, potentially struggling with ultra-short 1 ms predictions. Future research should explore specialized architectures or attention mechanisms for capturing patterns in these brief intervals.

 \item Transformers have high computational overhead, which can be inefficient for very short-term predictions. Future work should focus on model compression or lightweight variants to maintain accuracy while reducing computational demands. These efforts are essential for enhancing transformer-based models like FocalGatingNet for tasks demanding precise, low-latency predictions at extremely short timescales.
\end{itemize}




\section{CONCLUSIONS}
This study introduced FocalGatedNet, a novel deep learning model that enhances the Transformer architecture with Dynamic Contextual Focus Attention and Gated Linear Units. Our extensive experiments on a significant dataset demonstrate FocalGatedNet's superiority over several state-of-the-art models in predicting sagittal knee angles.
Notably, FocalGatedNet excels in longer prediction horizons, achieving an impressive 36\%  reduction in MAPE, emphasizing its ability to capture complex knee joint angle patterns.
Moreover, FocalGatedNet maintains a lower computational load, making it practical for real-time applications.
These findings highlight the significance of our model's enhancements, especially in long-term knee joint angle prediction. FocalGatedNet's results hold great promise for future research in wearable exoskeletons and neurological impairment rehabilitation.


\section*{Acknowledgment}
This work was supported in part by the Advanced Research and Innovation Center (ARIC), which is jointly funded by Mubadala UAE Clusters and Khalifa University of Science and Technology, and in part by Khalifa University Center for Autonomous and Robotic Systems under Award RC1-2018-KUCARS.


\bibliographystyle{chicago}
\bibliography{reference.bib}

\end{document}